%% file: PID4549.tex
\documentclass[letterpaper]{article}
\usepackage{aaai25}
\usepackage[utf8]{inputenc}
\usepackage[hyphens]{url}
\usepackage[table, xcdraw]{xcolor}
\usepackage{textcomp}
\usepackage{times}
\usepackage{helvet}
\usepackage{courier}
\usepackage{graphicx}
\usepackage{caption}
\usepackage{subcaption}
\usepackage{verbatim}
\usepackage{multirow}
\usepackage{tabularx}
\usepackage{booktabs}
\usepackage{makecell}
\usepackage{fdsymbol}
\usepackage{todonotes}
\usepackage{appendix}
\usepackage{tikz}
\usepackage{tikz-dependency}
\usepackage{listings}
\usepackage{newfloat}
\usepackage{amsfonts}
\usepackage{amsmath}
\usepackage{amsthm}
\usepackage{mathtools}
\usepackage{microtype}
\usepackage{latexsym}
\usepackage{color}
\usepackage{gensymb}
\usepackage{natbib}
\usepackage{algorithm}
\usepackage{algpseudocode}

\DeclareCaptionStyle{ruled}{labelfont=normalfont,labelsep=colon,strut=off}
\floatstyle{ruled}
\newfloat{listing}{tb}{lst}
\floatname{listing}{Listing}

\newcommand{\method}[1]{\ifthenelse{\equal{#1}{full}}{Confidence-Aware Defense}{CAD}}

\newcommand{\best}[1]{\textbf{#1}}

\graphicspath{{./figs/}{./figs/result/}}

\title{Mitigating Malicious Attacks in Federated Learning via Confidence-aware Defense}

\author{
    Qilei Li \textsuperscript{\rm 1},
    Pantelis Papageorgiou \textsuperscript{\rm 2},
    Gaoyang Liu \textsuperscript{\rm 3}, \\
    Mingliang Gao \textsuperscript{\rm 4}, 
    Chen Wang \textsuperscript{\rm 3},
    Ahmed M. Abdelmoniem \textsuperscript{\rm 1},
}

\affiliations{
    \textsuperscript{\rm 1}
    Queen Mary University of London,
    \quad \texttt{\{q.li, ahmed.sayed\}@qmul.ac.uk}, \\
    \textsuperscript{\rm 2} 
    University of Oxford, 
    \quad \texttt{pantelis.papageorgiou@cs.ox.ac.uk}, \\
    \textsuperscript{\rm 3} 
    Huazhong University of Science and Technology,
    \quad \texttt{\{liugaoyang, chenwang\}@hust.edu.cn}, \\
    \textsuperscript{\rm 4} 
    Shandong University of Technology,
    \quad \texttt{mlgao@sdut.edu.cn},
}

\graphicspath{{./figs/}{./figs/result/}}

\def\ie{\emph{i.e.}, }

\def\etal{\emph{et al.}}
\usepackage{bibentry}
\begin{document}

\maketitle

\begin{abstract}
    Federated Learning (FL) is an distributed machine learning diagram
    that enables multiple clients
    to collaboratively train a global model
    without sharing their private local data.
    However,
    FL systems are vulnerable to attacks that are happening
    in malicious clients
    through data poisoning and model poisoning
    who can deteriorate the aggregated global model’s performance.
    Existing defense methods typically focus on mitigating
    specific types of poisoning,
    and are often ineffective against unseen types of attack.
    These methods also assume an attack happened moderately
    while is not always holds true in real.
    Consequently,
    these methods can significantly fail
    in terms of accuracy and robustness
    when detecting and addressing updates
    from attacked malicious clients.
    To overcome these challenges,
    in this work,
    we propose a simple yet effective framework to detect malicious clients,
    namely \method{full}~(\method{abbr}),
    that utilizes the confidence scores
    of local models as criteria
    to evaluate the reliability of local updates.
    Our key insight is that malicious attacks,
    regardless of attack type,
    will cause the model to deviate from its previous state,
    thus leading to increased uncertainty when making predictions.
    Therefore,
    \method{abbr} is comprehensively effective
    for both model poisoning and data poisoning attacks
    by accurately identifying and mitigating potential malicious updates,
    even under varying degrees of attacks and data heterogeneity.
    Experimental results demonstrate
    that our method significantly
    enhances the robustness of FL systems against
    various types of attacks across various scenarios
    by achieving higher model accuracy
    and stability.

\end{abstract}

\section{Introduction}
\begin{figure}[ht]
    \centering
    \includegraphics[width=\linewidth]{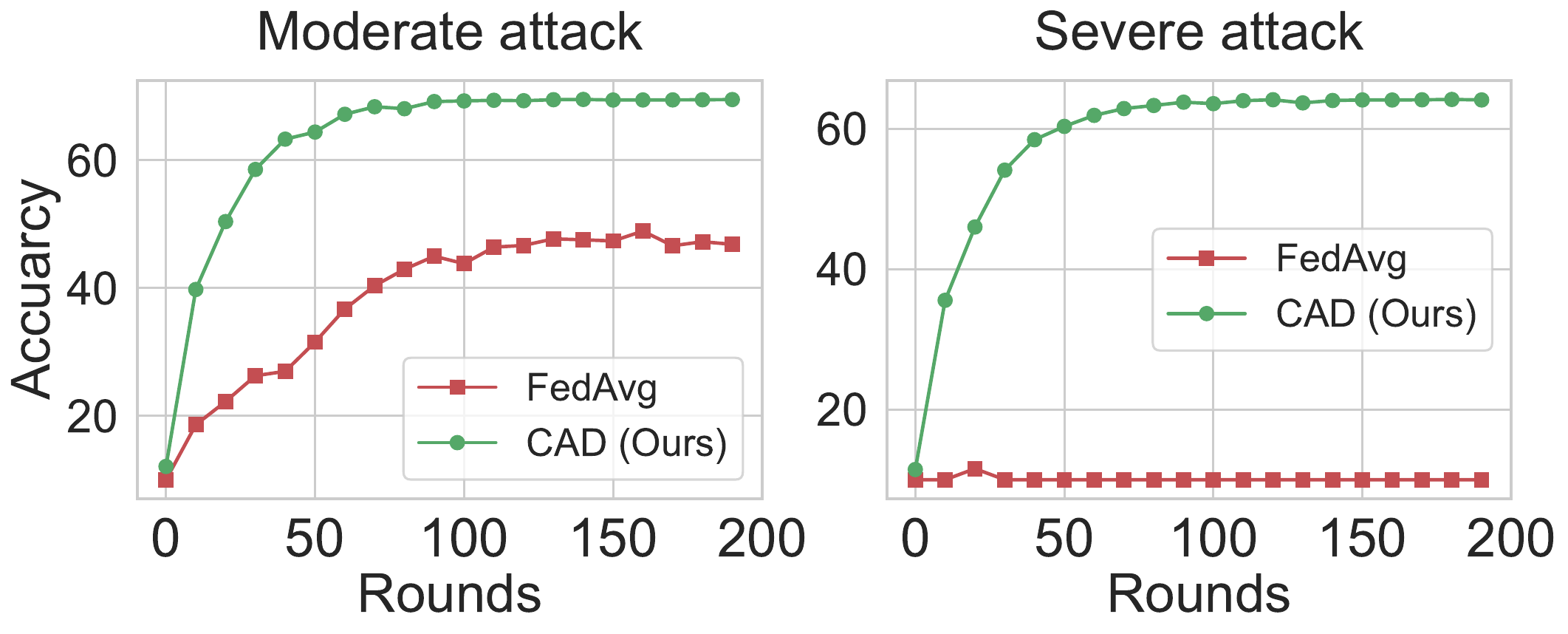}
    \caption{Impact of malicious attacks on model convergence
        under the ``Little Is Enough" attack at various intensities.
        The FedAvg aggregation strategy degrades significantly,
        while our \method{abbr} consistently maintains high accuracy and robustness.}
    \label{fig:attack}
\end{figure}

Federated Learning (FL)~\cite{zhang2021survey} is a distributed machine learning paradigm
that allows multiple clients
to collaboratively train a global model
without sharing their local data.
FL offers significant privacy protection,
therefore making it highly attractive
for privacy-sensitive applications~\cite{rodriguez2023survey}.
However,
the decentralized nature of FL also introduces security challenges:
malicious attacks happened in distributed local clients
can degrade the global model's performance
through data poisoning and model poisoning attacks,
leading to degraded predictions
and potentially catastrophic consequences~\cite{fang2023vulnerability, kumar2023impact}.
As illustrated in Fig~\ref{fig:attack},
malicious attacks
can significantly hinder the convergence of DNN model
and even completely cause collapse.
Therefore,
developing effective methods to detect and defend against malicious attacks
is crucial in ensuring the security and robustness of FL systems.

There are two main types of attacks in a FL system:
model poisoning and data poisoning,
categorized by where the attack occurs.
Model poisoning attacks involve directly manipulating the model updates
sent by the clients to the central server.
Among these,
Byzantine attacks~\cite{shi2022challenges}
are the most representative,
where malicious clients intentionally
send arbitrary or random updates
to disrupt the learning process
and significantly degrade the performance
of the global model~\cite{fang2020local, bhagoji2019analyzing}.
In contrast,
data poisoning attacks occur
when malicious clients intentionally
corrupt their local training data
to mislead the global model
during the training process,
such as label shuffle,
where the data-label pairs are randomly reassigned
that can confuse the globally updated local models~\cite{tolpegin2020data, xie2020dba}.
Both attack types pose significant threats,
and cause the global model to make incorrect predictions,
degrading its accuracy and robustness.
These challenges underscore the urgent need
for effective detection and defense mechanisms
capable of addressing various attack types.

Existing defense mechanisms primarily target
specific types of poisoning attacks,
under a moderate attack.
Byzantine-robust methods
such as Krum~\cite{blanchard2017machine}
and Trimmed Mean~\cite{yin2018byzantine}
aim to mitigate the impact of outliers
among client updates but often fall short
when dealing with data poisoning attacks.
However,
these methods generally assume that the majority of clients are honest,
which may not hold true in practical scenarios
where the proportion of malicious clients is significant~\cite{fang2020local}.
Recent advanced methods are primarily
driven by the analysis of fine-grained model consistency.
In that context,
FLDetector~\cite{zhang2022fldetector} improves upon existing defenses by detecting malicious clients through the consistency of their model updates across historical iterations,
and flags clients whose updates deviate significantly.
DeFL~\cite{yan2023defl} focuses on critical learning periods
within the training process
to determine the gradient changes and remove malicious clients.
FedRoLA~\cite{yan2024fedrola} enhances robustness by analyzing the per-layer consistency
of model updates
and conducts aggregation
based on the alignment of updates at each layer.
However, these methods overlook the uncertainty in model updates
and fail to address various types of both poisoning attacks comprehensively.

Given the inherent uncertainty of FL systems
and the attacks' complexity,
being capable of addressing only certain types of attack
is far from sufficient.
A robust and comprehensive defense approach
is needed to ensure the integrity
and reliability
of the global model
across all scenarios,
including various types of attacks
with differing levels of severity
and degrees of data heterogeneity.
To meet this demand,
we propose a simple yet effective defense model,
termed \method{full}~(\method{abbr}),
based on the confidence scores of clients' local models,
which measures how certain a model is about its predictions,
with higher scores indicating greater certainty
and lower scores reflect higher uncertainty.
Our key insight is that malicious attacks,
regardless of attack type,
will cause the model to deviate from its previous state,
thus leading to increased uncertainty in its predictions.
Therefore,
\method{abbr} is comprehensively effective
for both model poisoning and data poisoning attacks.
Specifically, the proposed approach includes the following steps:
1. Collecting confidence scores of each client update:
During each training round,
we collect the confidence scores of model updates from each client.
2. Establishing confidence boundaries:
Based on the collected confidence scores,
we assess the uncertainty of each client model update
and set the boundaries as a reference.
3. Detecting and handling malicious updates:
Based on the boundaries,
we identify updates with high uncertainty
and take appropriate actions,
such as discarding or adjusting these updates.

Our key contributions are summarized as follows:
\begin{itemize}
    \item We introduce a simple yet effective confidence-based
          malicious client detection method
          to identify if any attack occurs in a local client.
          Our approach is versatile and applicable to both
          model poisoning and data poisoning attacks,
          under various data heterogeneities.
    \item By detecting malicious attacks on client model updates
          through confidence scores,
          our method significantly improves the
          robustness of FL systems against attacks of different intensities.
          Whether facing moderate (25\% malicious clients),
          severe (50\% malicious clients),
          or extreme (75\% malicious clients) activities,
          our approach ensures higher model accuracy and stability.
    \item We validate the proposed method through extensive experiments
          on multiple datasets
          and demonstrate its effectiveness in enhancing the security and performance
          of FL models against a wide range of attacks.
          The results show that our confidence-based defense mechanism
          outperforms existing methods in terms of robustness and accuracy,
          in detecting and mitigating malicious updates.
\end{itemize}

\section{Related Work}
\subsection{Federated Learning (FL)}
FL is a decentralized machine learning diagram
that enables multiple clients to
collaboratively train a shared global model
while keeping their local data private.
In FL, the learning process is divided into two main stages:
local training and global aggregation.
During the local training phase,
each client trains a local model using its private dataset.
This involves updating the model parameters based on the local data,
which protects privacy by not allowing raw data to leave the client's device.
Once the local training is complete,
the clients send their updated model parameters to a central server for global aggregation,
in which the central server aggregates
the received model updates from participating clients
to form a new global model.
This aggregation can be done by various approaches,
the most common is the Federated Averaging (FedAvg) algorithm~\cite{mcmahan2017communication}.
The FedAvg algorithm computes the weighted average
of the model parameters from all clients
to form a global model.
Despite the advantages of FL in preserving data privacy,
the decentralized nature of this learning paradigm
makes it vulnerable to various types of attacks
either caused by data or model poisoning.
Therefore,
it is crucial to develop efficient and robust
defense mechanisms that can identify
and mitigate the impact of malicious activities.

\subsection{Malicious Client Attacks}
FL models are susceptible to various types of attacks
from malicious clients,
which can be broadly categorized into
data poisoning and model poisoning.

\vspace{0.5em}
\noindent
\textbf{Data Poisoning Attacks}: In data poisoning attacks,
malicious clients intentionally manipulate
their local training data to degrade the performance
of the global model.
Common strategies include label flipping,
where the labels of training samples are deliberately swapped,
and label shuffling,
where the labels of training samples are randomly reassigned.
These attacks aim to bias the model's training process,
and further lead to inaccurate or harmful predictions.
Rahman \etal~\citet{rahman2023federated} highlights
how local model poisoning through data manipulation
can significantly impact the robustness
of federated learning systems.
Similarly, Smith \etal~\cite{smith2023defense} examine clean-label poisoning attacks,
where the manipulated data appears legitimate but is designed to subvert the performance of the model.

\vspace{0.5em}
\noindent
\textbf{Model Poisoning Attacks}:
In contrast to data poisoning attacks,
model poisoning attacks refer to malicious clients
directly sending manipulated model updates
to the central server during the global aggregation phase.
These attacks are conducted by altering the model parameters
in a way
that maximizes the negative impact on the global model.
Zhang \etal~\cite{zhang2023backdoor} have demonstrated how backdoor attacks could be performed by injecting specific patterns into the model updates, enabling attackers to trigger incorrect behaviors under certain conditions. Lee \etal~\cite{lee2023adversarial} analyze the adversarial impact of model attacks on federated learning models.
Additionally,
Byzantine attacks,
which involve arbitrary or adversarial behavior by clients,
can also be classified under model poisoning
when clients send random or adversarial updates to the central server.
Recent Byzantine-robust methods like Krum++~\cite{wang2023krumpp}
and Enhanced Trimmed Mean~\cite{liu2023enhanced}
aim to filter out these abnormal updates
but can struggle when the attacks are sophisticated.

\subsection{Defense Against Malicious Attacks}
To detect and mitigate the aforementioned attacks, numerous works have been proposed recently~\cite{fang2023vulnerability,yan2023defl,zhang2022fldetector, yan2024fedrola}.
DeFL~\cite{yan2023defl} is proposed as a defense mechanism
that secures critical learning periods during the training process
to reduce the impact of model poisoning attacks.
Zhang \etal~\cite{zhang2022fldetector} introduced FLDetector to integrate statistical analysis, anomaly detection,
and machine learning methods to enhance the accuracy of detecting poisoned updates.
FedRoLA~\cite{yan2024fedrola} strengthens robustness by analyzing per-layer consistency in model updates and conducting aggregation based on the alignment of updates at each layer.
Despite the improved accuracy in detecting malicious clients, these existing methods are often tailored to specific types of attacks. Additionally, they typically rely on the fragile assumption that the attack intensity is moderate,
meaning less than 25\% of the participating clients are malicious,
which may not hold in a practical FL system where higher proportions of malicious clients can be present.
To resolve these limitations,
in this paper,
we propose the \method{abbr} model
to address both types of attacks and varying intensities.

\subsection{Model Confidence and Uncertainty}
Confidence scores~\cite{gass1981concepts} measure a model's certainty
regarding its predictions.
Higher confidence scores indicate greater confidence,
while lower scores suggest higher uncertainty.
It provides insights into the model's decision-making process
by identifying the predictions
that the model is less sure about~\cite{gal2016dropout,lakshminarayanan2017simple}.

\vspace{0.5em}
\noindent
\textbf{Super Loss}
is a confidence-aware loss designed for curriculum learning,
which incorporates confidence scores to determine the reliability
of each training instance~\cite{castells2020superloss}.
it introduces a regularization
to adjust the loss dynamically
based on the confidence score (${\sigma}^j$) for sample $j$
in the training set $\mathcal{D}$.
The Super Loss is defined as:
\begin{equation}
    \mathcal{L}_{\text{super}} = \frac{1}{N} \sum_{j=1}^{N} \left( \frac{1}{{\sigma^j}^2} \mathcal{L}_j + \log \sigma_j \right),
\end{equation}
where $\mathcal{L}_i$ is the task loss for sample $i$,
and ${\sigma}^j$ is the confidence score,
and $N$ is the number of samples.
The confidence score ${\sigma}^j$ is obtained by the following closed-form solution:
\begin{equation}
    {\sigma}^j = \exp \left( -\text{W} \left( \frac{1}{2} \max \left( -\frac{2}{e},
            \frac{\mathcal{L}_i - \log(C)}{\lambda} \right) \right) \right),
    \label{equ:sigma}
\end{equation}
where $\log(C)$ is the constant,
and $C$ is set to the number of classes,
and $\lambda$ is the regularization parameter.
In this work,
motivated by the observation that
malicious attack will disrupt the local model's
convergence against the original global initialization,
and making the learned model less confident in the output,
we propose to use per-client confidence accumulation
as the index to determine the quality of the learned model.
Note that we use Eq.~\eqref{equ:sigma} solely for estimating the confidence score,
in contrast to the original design of Super Loss which uses it as a regularization term.
The rationale is that our goal is to make the \method{abbr} model
versatile and applicable to various tasks,
and to explicitly demonstrate that the effectiveness
stems from the detection algorithm
rather than modifications to the loss function.

\section{Methodology}

\begin{figure*}[ht]
    \centering
    \includegraphics[width=\linewidth]{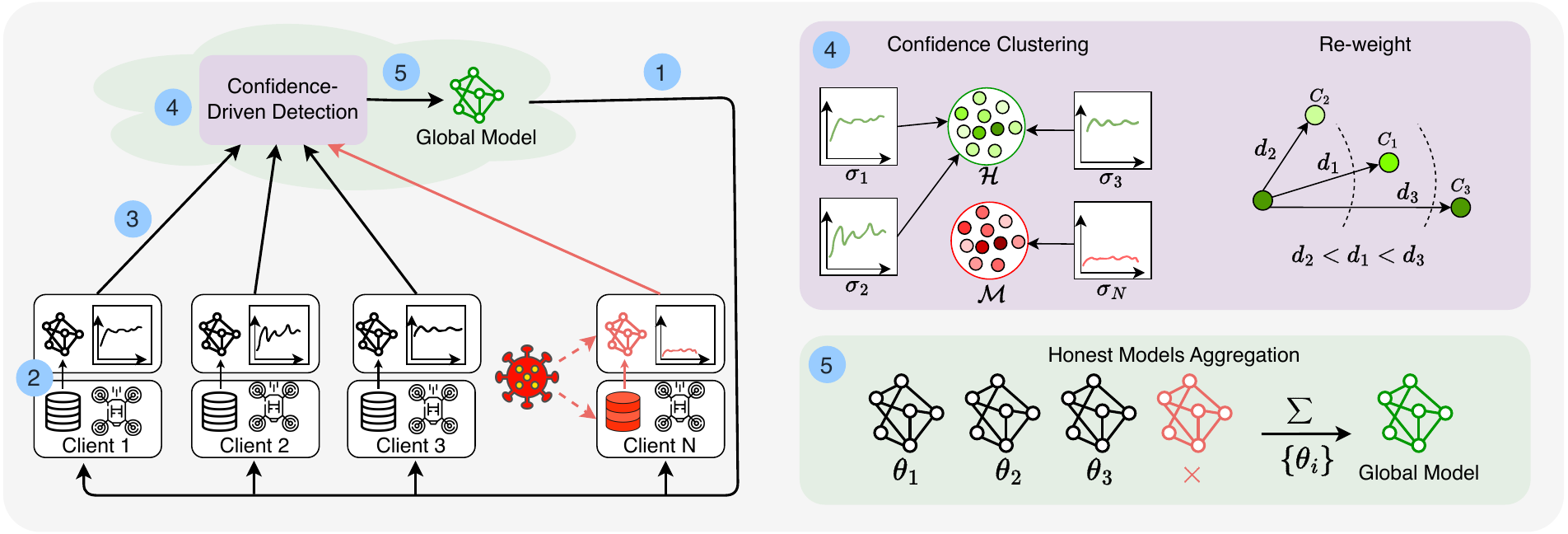}
    \caption{
        The proposed \protect \method{full} (\method{abbr}) framework.
        It encompasses five key steps:
        1. Initialize the global model and distribute it to clients.
        2. Clients train their personalized, confidence-aware local models.
        3. Clients upload their local models and associated confidence scores to the server.
        4. The server identifies potentially malicious clients through confidence clustering.
        5. The server aggregates the local models of honest clients
        by re-weighted aggregation.
    }
    \label{fig:framework}
\end{figure*}

\subsection{Problem Formulation}
We consider an FL system with $n$ clients,
each with its local dataset $\mathcal{D}_i$.
The goal is to train a global model $\theta$
by aggregating the local updates $\theta_i$ from client $\{C_i\}_{i=1}^n$
while mitigating the impact caused by potential malicious clients,
which can perform two types of attacks:
\textit{\textbf{Data poisoning attacks}}:
These include label manipulation
where the local dataset is intentionally corrupted to mislead the global model.
\textit{\textbf{Model poisoning attacks}}:
Here, the model updates sent by the clients are manipulated
to degrade the performance of the global model.
To achieve this goal,
we propose a \method{full} (\method{abbr}) mechanism
based on the per-client confidence scores ($\{\sigma_i\}_{i=1}^n$),
which reflects
the uncertainty of the local model's predictions
and helps in identifying malicious updates.
The notations used in formulating \method{abbr} are summarized in Table~\ref{tab:notations}.

\begin{table}[ht]
    \centering
    \caption{Table of notations used in this paper.}
    \input{tabs/notation}
    \label{tab:notations}
\end{table}

\subsection{Confidence-Driven Detection}

\subsubsection{Client Confidence Estimation.}
For each client,
the confidence score is estimated
with their local dataset $\mathcal{D}_i$.
Referring to Eq.~\eqref{equ:sigma},
for each local client,
we calculate the accumulated confidence score for
all samples in the local dataset
and then compute an average
to obtain the per-client confidence score $\sigma_i$.
by averaging the accumulated confidence scores
for client $C_i$,
which is the average confidence scores
among all samples in its dataset:
\begin{equation}
    \sigma_i  = \frac{1}{N_i} \sum_{j=1}^{N_i} \sigma^j_i,
\end{equation}
where $N_i$ is the number of samples in $\mathcal{D}_i$.
This $\sigma_i$ reflects
the overall uncertainty of the local model's predictions
for client $i$.
Once the confidence scores for all clients have been computed,
they will be normalized
to ensure comparability
across different clients
by min-max scaling,
which transforms the confidence scores into a uniform range [0, 1].
Let $S = \{\sigma_i\}_{i=1}^{n}$ represent the set of confidence scores
from $n$ clients.
The normalized confidence scores
$S_{\text{norm}}$ are formulated as:
\begin{equation}
    S_{\text{norm}} = \frac{S - \min(S)}{\max(S) - \min(S)}.
\end{equation}

\subsubsection{Clustering Based Detection.}
After normalizing the per-client confidence scores,
the next step is to cluster the clients into groups
of honest and potentially malicious clients.
The rationale for such clustering
is based on the assumption that
clients can be divided into two groups:
honest clients with higher confidence scores
and potentially malicious clients with lower confidence scores.
To achieve this,
we apply two-way k-means clustering
to the normalized confidence scores
$S_{\text{norm}}$.
By setting the number of clusters to 2,
we obtain two centroids
to establish the lower bound and upper bound:
\begin{equation}
    \{\mu_{\text{lower}}, \mu_{\text{upper}}\} = \text{k-means}(S_{\text{norm}}, 2).
\end{equation}
Once the centroids are obtained,
the following step is to classify the clients as honest or malicious
based on the distance
in terms of the normalized confidence score
to the cluster centroids.
Specifically,
a client $i$ is deemed honest
if its normalized confidence score $\sigma_i$
is closer to the centroid $\mu_{\text{upper}}$ (representing higher confidence)
than to the centroid $\mu_{\text{lower}}$ (representing lower confidence).
Conversely, a client is classified as malicious if the opposite holds true:
\begin{equation}
    \begin{aligned}
        \mathcal{H} & = \{i \mid |\sigma_i - \mu_{\text{upper}}| < |\sigma_i - \mu_{\text{lower}}|\},    \\
        \mathcal{M} & = \{i \mid |\sigma_i - \mu_{\text{lower}}| \leq |\sigma_i - \mu_{\text{upper}}|\}.
    \end{aligned}
\end{equation}
By classifying clients in this manner,
we can focus on aggregating updates from only honest clients
and get rid of malicious ones,
thereby enhancing the robustness and security of an FL system.

\subsection{Re-weighted Aggregation}
Based on the identified set of malicious clients,
the next step is to perform weight aggregation.
Let $\Theta_{\mathcal{H}}$ denote the set of honest local model updates,
and $\mathcal{L}_{\mathcal{H}}$ represent
the corresponding set of data lengths for these honest clients.
These are defined as:
\begin{equation}
    \Theta_{\mathcal{H}} = \{\theta_i \mid i \in \mathcal{H}\},
\end{equation}
\begin{equation}
    \mathcal{L}_{\mathcal{H}} = \{\ell_i \mid i \in \mathcal{H}\},
\end{equation}
by which we isolate the local updates of clients
that are considered trustworthy,
so as to ensure that only reliable information
will be used in the aggregation process.
Subsequently,
we define the original weights $w_{\text{orig},i}$
based on the data lengths of the honest clients.
The original weight for each client
is calculated as the ratio of the client's data
length to the total data length of all honest clients:
\begin{equation}
    W_{\text{orig}}  = \{w_{\text{orig},i}\}_{i \in \mathcal{H}} = \{\frac{\ell_i}{\sum_{j \in \mathcal{H}} \ell_j}\}_{i \in \mathcal{H}}.
\end{equation}
In addition,
to \textit{adaptively} optimize the weight
based on the trustworthiness (confidence)
of each honest client,
we apply a re-weighting strategy,
for which a re-weighting regularization factor $r_i$ is then calculated
for each honest client
based on their normalized confidence scores.
The set of re-weighting regularization factors for honest clients is denoted as:
\begin{equation}
    R = \{\sigma_i \mid i \in \mathcal{H}\}.
\end{equation}
The factors are then normalized:
\begin{equation}
    r_{\text{norm},i} = \frac{\sigma_i}{\sum_{j \in \mathcal{H}} \sigma_j},
\end{equation}
so that
one can ensure
that the aggregation process
give appropriate importance to individual client models.
Once calculated the original weights
and the re-weighting regularization factors,
we combine both to obtain the final weights as:
\begin{equation}
    w_{\text{final},i} = r_{\text{norm},i} \cdot  w_{\text{orig},i}.
    \label{eq:reweight}
\end{equation}
As a final step,
the global model parameters are then aggregated
by the re-weighted average,
which ensures the contributions
from different clients are proportional
to their combined confidence scores
and original weights.
Let $\Theta_{\mathcal{H}}$ be the set of models from the honest clients,
then the updated global model $\theta_{\text{global}}$ is obtained as:
\begin{equation}
    \theta_{\text{global}} = \frac{\sum_{i \in \mathcal{H}} w_{\text{final},i} \cdot \theta_i \cdot \ell_i}{\sum_{i \in \mathcal{H}} w_{\text{final},i} \cdot \ell_i}.
\end{equation}

By interactively conducting this detection and selected aggregation,
our proposed \method{abbr} model effectively identifies
and aggregates updates from honest clients,
while ensuring that the aggregation process gives more weight to clients
with higher confidence scores,
thus improving the overall performance and security of the FL model.

\section{Experiments}
\subsection{Experimental Setup}

\begin{table*}[ht]
    \centering
    \caption{Robustness to various intensities of attack on CIFAR-10 with VGG-11 model.
        Best results are shown in \textbf{bold}.}
    \input{tabs/sota_level.tex}
    \label{tab:sota_level}
\end{table*}

\subsubsection{Datasets.}
We evaluated \method{abbr} model a wide range of benchmarks including:
CIFAR-10~\cite{krizhevsky2009learning},
MNIST~\cite{lecun1998gradient},
and Fashion-MNIST~\cite{xiao2017fashion} datasets,
which are common in prior studies~\cite{yan2023defl,yan2024fedrola}.
We employed a heterogeneous partitioning strategy
to create non-IID local data
where samples and class distributions
are unbalanced among clients.
The factor is sampled from a Dirichlet distribution
$\texttt{Dir}_N(\alpha)$,
with $\alpha$ determining the level of heterogeneity,
where a smaller $\alpha$ value refers
to high heterogeneity~\cite{shang2022federated}.
We set $\alpha = 0.5$ by default
in the following experiments.
We evaluated the performance
by classification accuracy.

\subsubsection{DNN Models.}
We tested \method{abbr} model on three representative DNN models including:
AlexNet~\cite{krizhevsky2012imagenet},
VGG-11~\cite{simonyan2015very},
and a Multi-Layer Perceptron (MLP)~\cite{popescu2009multilayer}
with layers consisting of 784, 512, and 10 neurons respectively.
Specifically,
AlexNet and VGG-11 were employed for the CIFAR-10 dataset,
the MLP for MNIST,
and AlexNet for Fashion-MNIST.

\subsubsection{Baseline Defenses.}
We compared \method{abbr} with state-of-the-art defense models including:
TrimMean~\cite{yin2018byzantine},
AFA~\cite{munoz2019byzantine},
FLDetector~\cite{zhang2022fldetector},
DeFL~\cite{yan2023defl},
FedRoLA~\cite{yan2024fedrola}.
Additionally,
we included FedAvg~\cite{mcmahan2017communication}
as a baseline method for comparison.
We consider two settings regarding the adversary’s knowledge~\cite{yan2023defl}:
(a) \textit{\textbf{Full}},
where the adversary knows the gradients of honest clients,
and (b) \textit{\textbf{Partial}},
where the adversary is unaware of the gradient updates shared by honest clients.
We evaluated the detection performance
with True Positive Rate (TPR)
and False Positive Rate (FPR).

\subsubsection{Malicious Client Attacks.}
We tested \method{abbr} model on
both model poisoning attacks and data poisoning attacks:

\noindent\textbullet\
\textbf{Label Shuffling (LS)}:
Malicious clients randomly
reassign labels of their local data,
creating noisy and misleading training data
that reduces the global model's accuracy.

\noindent\textbullet\
\textbf{Little is Enough (LIE)}:
Adds small amounts of noise
to the average of honest gradients.
The adversary computes the average $\mu$
and standard deviation $\sigma$ of honest gradients,
then applies a coefficient $z$
to obtain the malicious update $\mu + z\sigma$.

\noindent\textbullet\
\textbf{Min-Max (MM)}:
Computes malicious gradients
such that their maximum distance from any gradient
is upper bounded by the maximum distance
between any two honest gradients,
placing them close to the honest gradient cluster.

\noindent\textbullet\
\textbf{Min-Sum (MS)}:
The sum of squared distances of malicious gradients
from all honest gradients is upper bounded
by the sum of squared distances among honest gradients.
Malicious gradients are kept identical to maximize impact.

\subsubsection{Implementation Details.}
We simulated $N = 50$ clients in the experiments.
Each client was trained for 20 epochs
with stochastic gradient descent to update its local model.
The number of training rounds was set to 200
to ensure convergence for all DNN models.
The learning rate was set to 0.01.
The batch size was set to 16,
with a weight decay of $1 \times 10^{-3}$.
We evaluated the robustness
of our \method{abbr} model
under three degrees of attack severity
defined by the percentage of malicious clients:
\textit{moderate} (25\%),
\textit{severe} (50\%),
and \textit{extreme} (75\%).
For data poisoning attacks,
confidence scores were directly obtained
during model training.
For model poisoning attacks,
confidence scores were obtained
by further running inference on local training data
before sending it to the parameter server.
We implemented our defenses,
in PyTorch~\cite{paszke2019pytorch} on a single NVIDIA A100 GPU.

\subsection{Experimental Results}

\subsubsection{Comparison on Defense Robustness
    under Different Attack Intensities.}
We evaluated the robustness of the proposed \method{abbr} model
and other state-of-the-art defenses
under different levels of attack intensity
with the CIFAR-10 dataset with the VGG model.
The results, as shown in Table~\ref{tab:sota_level},
indicate that under moderate attacks (25\% malicious clients),
most defense methods perform reasonably well,
with our \method{abbr} model
achieving the highest accuracy
across all attack types.
However, as the intensity of the attacks
increases to severe (50\%) and extreme (75\%),
the performance of most methods significantly deteriorates.
In contrast,
the \method{abbr} model consistently
maintains high accuracy and robustness
by outperforming other methods under all attack intensities.
This demonstrates the effectiveness of
our confidence-based detection
and re-weighted aggregation approach
in mitigating the impact of various attack types,
even under severe attack conditions.

\subsubsection{Applicable to Various Models and Datasets.}
Table~\ref{tab:sota_med} compares the performance
of \method{abbr} with other state-of-the-art defense methods
across different models and datasets.
The results demonstrate that \method{abbr} consistently outperforms
other defenses with high accuracy under various attacks.
For MNIST with MLP,
most defense methods perform well under label shuffling attacks,
but \method{abbr} achieves the highest accuracy
across all attack types.
Similarly,
for Fashion MNIST with AlexNet,
\method{abbr} maintains superior performance,
especially against LIE and MM attacks,
where other methods
significantly collapsed in accuracy.
In the CIFAR-10 dataset,
both with VGG-11 and AlexNet models,
the robustness of \method{abbr} stands out.
While other defenses struggle to converge,
particularly under LIE and MM attacks,
\method{abbr} consistently achieves the best results.
This highlights the effectiveness of our method
in providing robust defense against various attacks,
even under different model and dataset combinations.
These results demonstrated that \method{abbr}
is not only effective in mitigating attacks for a specific benchmark
but also adaptable to various models and datasets,
thereby making it a reliable defense method.

\begin{table}[ht]
    \centering
    \caption{Defense robustness across various models and datasets.
        Accuracy of different defense methods
        with various attacks
        across different models and datasets
        under the severe attack setting.
        The best results are shown in \textbf{bold}.}
    \input{tabs/sota_med.tex}
    \label{tab:sota_med}
\end{table}

\subsubsection{Robustness to Data Heterogeneity.}
\begin{figure}[ht]
    \centering
    \includegraphics[width=\linewidth]{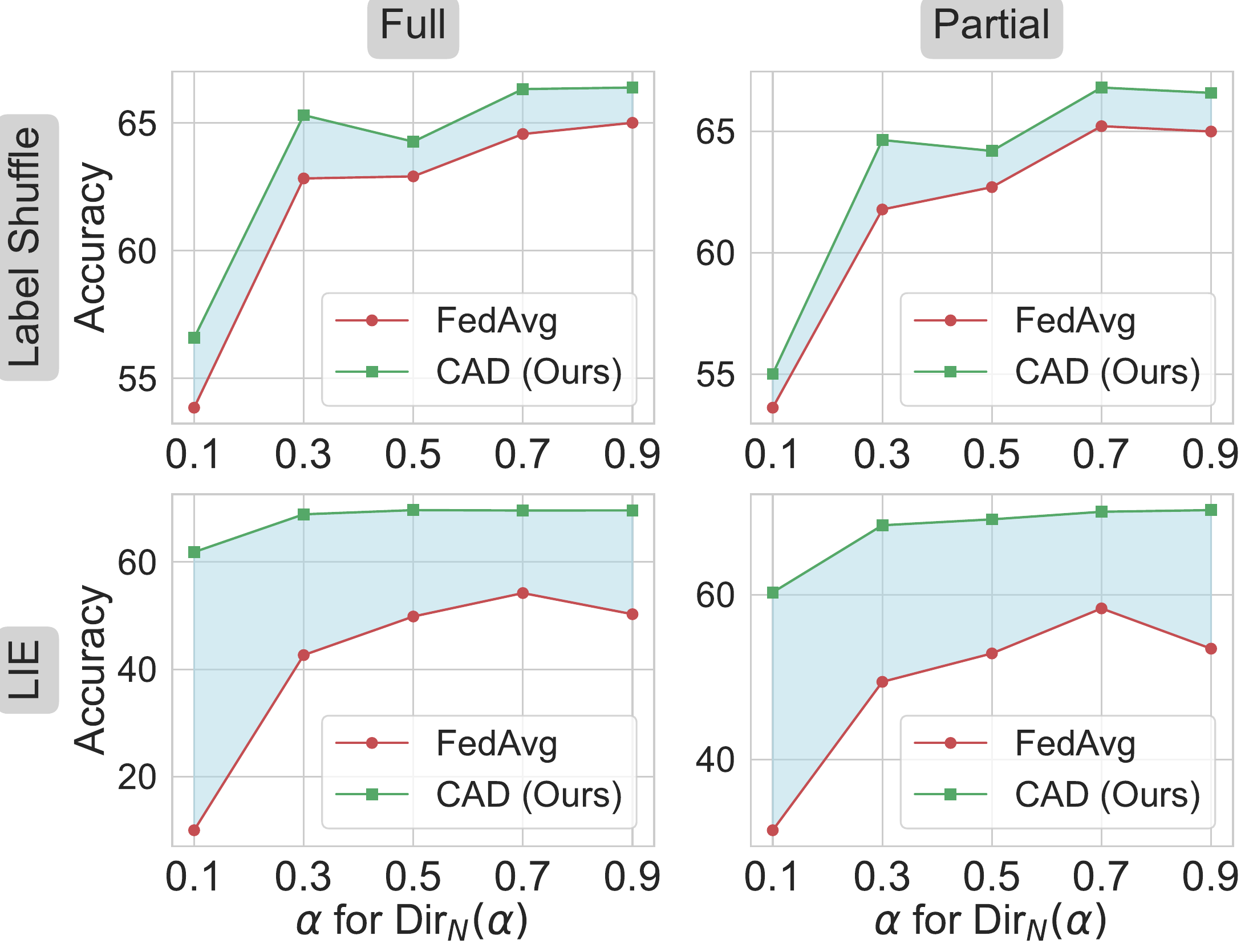}
    \caption{Robustness to data heterogeneity
        by varying degrees of Non-IID data,
        as controlled by the parameter $\alpha$ in the Dirichlet distribution $\texttt{Dir}(\alpha)$. }
    \label{fig:alpha}
\end{figure}
We further investigated the robustness of \method{abbr}
across different levels of data heterogeneity
by varying the key parameter $\alpha$
in the Dirichlet distribution $\text{Dir}(\alpha)$.
A smaller $\alpha$ indicates a higher degree of non-IID data.
We simulated a moderate LIE attack and a severe Label Shuffle
attack on the Cifar10 dataset with VGG-11 model.
As shown in Fig.~\ref{fig:alpha},
as the non-IID degree decreases (\ie, as $\alpha$ increases),
the global model accuracy improves at a much higher rate compared to FedAvg,
as indicated by the blue shaded area,
consistently under both Full and Partial scenarios.
These results confirm the robustness of our \method{abbr} model
to varying data heterogeneity.

\subsubsection{Malicious Clients Detection Accuracy.}
\begin{figure}[ht]
    \centering
    \includegraphics[width=\linewidth]{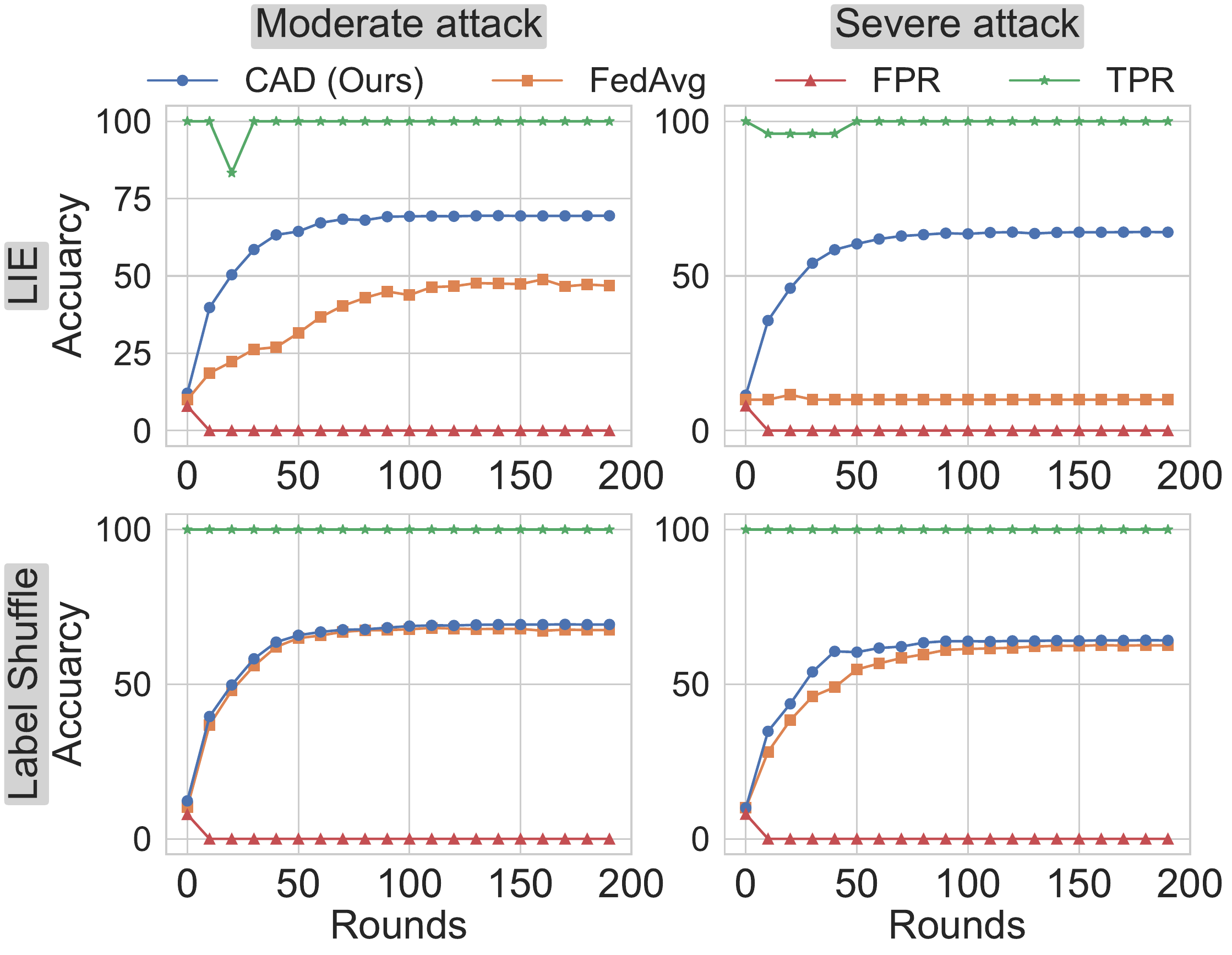}
    \caption{Malicious client detection accuracy evaluated on the Cifar10 dataset with the VGG-11 model.}
    \label{fig:detection_acc}
\end{figure}

We evaluated the detection performance
of our proposed \method{abbr} model
under LIE and label shuffle attack scenarios
considering both moderate and severe intensities.
The evaluation was conducted on the Cifar10 dataset with the VGG-11 model
at full attack setting.
As shown in Fig.~\ref{fig:detection_acc},
our \method{abbr} model can accurately detect malicious clients,
and converge to high TPR and low FPR
after only a few training rounds.
Therefore,
it can exclude malicious clients and lead to a significant improvement in overall performance,
as shown in the blue line over the orange line.
Its advantage is more obvious when
more clients are being attacked,
with detection accuracy remaining high and stable
even under severe attack conditions
while the FedAvg was collapsed.
This demonstrates that \method{abbr}
is effective under varying intensities
and thereby significantly enhances model performance compared to the FedAvg baseline.

\subsubsection{Impact of Re-weighting.}
\begin{figure}[h]
    \centering
    \includegraphics[width=\linewidth]{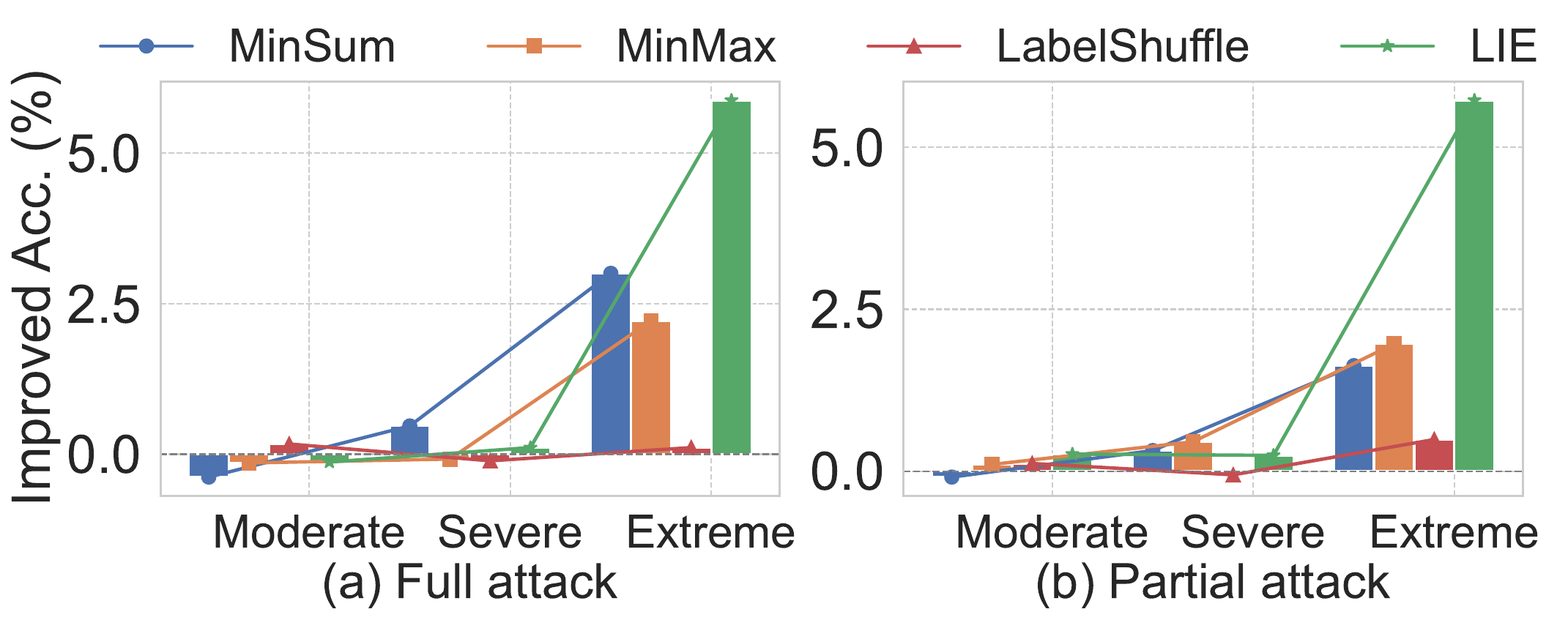}
    \caption{Effect of re-weighting on model accuracy
        across different degrees of attack.}
    \label{fig:reweight}
\end{figure}

We ablated the impact of the re-weighting operation in Eq.~\eqref{eq:reweight}
and the results are shown in Fig.~\ref{fig:reweight}.
It can be observed that
the re-weighting operation can consistently enhance performance,
especially at higher attack rates.
The improvement is particularly clear in ``Severe” and ``Extreme” conditions,
with detecting attacks like LIE and MinSum benefiting the most.
This effectiveness is likely due to the fact
that only a few honest clients remain under high attack rates,
therefore, it is crucial to identify and prioritize them.
This further demonstrates that the confidence score is a reliable metric
for assessing the quality of client models,
thereby supporting the underlying assumption
of our confidence-aware \method{abbr} model.

\section{Conclusion}
In this paper,
we proposed the \method{abbr} model,
an attack defense mechanism for FL
that achieves accurate malicious client detection
to address both model poisoning and data poisoning attacks.
The proposed \method{abbr} model estimates per-client confidence scores
to reflect the model's uncertainty,
which are then referred
to identify clients into honest and malicious categories.
By re-weighting the contributions of honest clients
based on their confidence scores,
\method{abbr} model can make the best use of honest clients
even under severe attack when there is only a handful of honest clients available,
so that it can enhance the robustness
of the global model.
Extensive experiments conducted on various benchmarks
with non-IID data demonstrated that \method{abbr} can
notably enhance the accuracy,
stability,
and robustness of FL systems.

\bibliography{PID4549}
\end{document}

%% file: tabs/notation.tex
\begin{tabular}{@{}ll@{}}
    \toprule
    \textbf{Notation}                        & \textbf{Description}                                        \\ \midrule
    $n$                                      & Number of clients                                           \\
    $\mathcal{D}_i$                          & Local dataset of client $i$                                 \\
    $\theta$                                 & Global model parameters                                     \\
    $\theta_i$                               & Local model parameters of client $i$                        \\
    $\mathcal{L}_i^j$                        & Individual loss for sample $j$ in $\mathcal{D}_i$           \\
    $\sigma_i$                               & Confidence score for client $i$                             \\
    $S$                                      & Set of confidence scores from all clients                   \\
    $\mu_{\text{lower}}, \mu_{\text{upper}}$ & Centroids from k-means clustering                           \\
    $\mathcal{H}$                            & Set of honest clients                                       \\
    $\mathcal{M}$                            & Set of malicious clients                                    \\
    $\mathcal{L}_{\mathcal{H}}$              & Data lengths of honest clients                              \\
    $W_{\text{orig}}$                        & Original weights based on data lengths                      \\
    $R$                                      & Re-weight regularization factors  \\
    $r_{\text{norm},i}$                      & Normalized regularization factor  \\
    $w_{\text{final},i}$                     & Final weight for client $i$                                 \\
    $\text{W}$                               & Lambert W function                                          \\ \bottomrule
\end{tabular}

%% file: tabs/sota_level.tex
\setlength{\tabcolsep}{12pt}
\begin{tabular}{clcccccccc}
    \toprule
    \multicolumn{1}{c}{\multirow{2}[2]{*}{\centering Intensity}}
                    & \multirow{2}[2]{*}{\centering Defence}
                    & \multicolumn{2}{c}{Min-Max}
                    & \multicolumn{2}{c}{Min-Sum}
                    & \multicolumn{2}{c}{LIE}
                    & \multicolumn{2}{c}{Label Shuffle}                                                                                                                              \\
    \cmidrule{3-10} &                                        & Full         & Partial      & Full         & Partial      & Full         & Partial      & Full         & Partial      \\
    \midrule
    \multirow{7}{*}{\makecell{Moderate                                                                                                                                               \\ (25\%)}}
                    & FedAvg                                 & 55.40        & 64.37        & 49.44        & 59.20        & 49.84        & 52.89        & 68.32        & 68.60        \\
                    & TrimMean                               & 53.69        & 61.35        & 51.58        & 60.33        & 31.63        & 34.22        & 65.20        & 65.20        \\
                    & AFA                                    & 68.89        & 68.71        & 68.99        & 69.17        & 69.09        & 68.80        & 68.88        & 69.09        \\
                    & FLDetector                             & 32.66        & 32.37        & 25.93        & 35.61        & 61.94        & 56.92        & 39.74        & 46.32        \\
                    & DeFL                                   & 65.67        & 69.04        & 62.97        & 68.74        & 48.02        & 51.51        & 67.95        & 68.57        \\
                    & FedRoLA                                & 69.23        & 69.03        & 68.59        & 68.76        & 68.67        & 68.95        & 68.48        & 68.59        \\
                    & \method{abbr} (Ours)                   & \best{69.85} & \best{69.43} & \best{69.94} & \best{69.43} & \best{69.67} & \best{69.18} & \best{69.36} & \best{69.47} \\
    \midrule
    \multirow{7}{*}{\makecell{Severe                                                                                                                                                 \\ (50\%)}}
                    & FedAvg                                 & 19.02        & 18.37        & 21.40        & 22.25        & 12.94        & 15.04        & 62.91        & 62.70        \\
                    & TrimMean                               & 16.92        & 17.27        & 17.49        & 17.20        & 11.55        & 12.35        & 37.26        & 36.88        \\
                    & AFA                                    & 16.65        & 17.89        & 15.42        & 17.42        & 10.42        & 10.56        & 60.32        & 60.85        \\
                    & FLDetector                             & 23.61        & 20.02        & 20.47        & 20.73        & 14.41        & 11.96        & 34.84        & 38.81        \\
                    & DeFL                                   & 21.60        & 27.50        & 48.26        & 48.26        & 23.17        & 23.95        & 63.02        & 62.31        \\
                    & FedRoLA                                & 19.52        & 19.60        & 21.40        & 22.10        & 13.42        & 14.17        & 62.52        & 61.98        \\
                    & \method{abbr} (Ours)                   & \best{64.00} & \best{65.07} & \best{64.75} & \best{65.13} & \best{64.44} & \best{64.96} & \best{64.27} & \best{64.19} \\
    \midrule
    \multirow{7}{*}{\makecell{Extreme                                                                                                                                                \\ (75\%)}}
                    & FedAvg                                 & 16.71        & 17.72        & 17.79        & 17.60        & 10.56        & 12.08        & 50.51        & 53.46        \\
                    & TrimMean                               & 14.57        & 15.03        & 17.52        & 16.60        & 10.01        & 10.42        & 21.17        & 20.23        \\
                    & AFA                                    & 11.54        & 11.62        & 13.84        & 10.03        & 10.06        & 10.51        & 20.23        & 22.17        \\
                    & FLDetector                             & 16.71        & 17.72        & 17.79        & 16.31        & 11.04        & 12.08        & 18.33        & 18.04        \\
                    & DeFL                                   & 20.11        & 19.44        & 19.42        & 20.02        & 16.06        & 16.76        & 52.18        & 52.72        \\
                    & FedRoLA                                & 16.71        & 16.86        & 17.79        & 17.59        & 10.00        & 10.00        & 54.39        & 52.46        \\
                    & \method{abbr} (Ours)                   & \best{56.96} & \best{57.04} & \best{56.88} & \best{56.48} & \best{56.75} & \best{56.54} & \best{57.68} & \best{57.79} \\
    \bottomrule
\end{tabular}%

%% file: tabs/sota_med.tex
\begin{tabular}{clcccc}
      \toprule
      Model & Defence              & \multicolumn{1}{c}{MM} & \multicolumn{1}{c}{MS} & \multicolumn{1}{c}{LIE} & \multicolumn{1}{c}{LS} \\
      \midrule
      \multirow{7}{*}{\makecell{Cifar10                                                                                                 \\ (VGG-11)}}
            & FedAvg               & 19.02                  & 21.40                  & 12.94                   & 62.91                  \\
            & AFA                  & 16.65                  & 15.42                  & 10.42                   & 60.32                  \\
            & FLDetector           & 23.61                  & 20.47                  & 14.41                   & 34.84                  \\
            & DeFL                 & 21.60                  & 48.26                  & 23.17                   & 63.02                  \\
            & FedRoLA              & 19.52                  & 21.40                  & 13.42                   & 62.52                  \\
            & \method{abbr} (Ours) & \best{64.00}           & \best{64.75}           & \best{64.44}            & \best{64.27}           \\
      \midrule
      \multirow{7}{*}{\makecell{Cifar10                                                                                                 \\ (AlexNet)}}
            & FedAvg               & 19.86                  & 19.75                  & 10.16                   & 59.63                  \\
            & AFA                  & 18.68                  & 19.47                  & 10.01                   & 57.02                  \\
            & FLDetector           & 17.24                  & 17.17                  & 10.00                   & 32.47                  \\
            & DeFL                 & 30.69                  & 49.05                  & 18.98                   & 59.42                  \\
            & FedRoLA              & 19.52                  & 19.75                  & 10.16                   & 59.33                  \\
            & \method{abbr} (Ours) & \best{62.92}           & \best{62.60}           & \best{62.90}            & \best{62.76}           \\
      \midrule
      \multirow{7}{*}{\makecell{Fashion                                                                                                 \\ MNIST \\ (AlexNet)}}
            & FedAvg               & 78.71                  & 74.58                  & 15.66                   & 88.06                  \\
            & AFA                  & 23.70                  & 23.70                  & 10.00                   & 88.56                  \\
            & FLDetector           & 78.37                  & 74.98                  & 10.00                   & 76.51                  \\
            & DeFL                 & 87.73                  & 86.83                  & 10.11                   & 88.25                  \\
            & FedRoLA              & 78.09                  & 74.58                  & 15.66                   & 88.05                  \\
            & \method{abbr} (Ours) & \best{88.43}           & \best{88.32}           & \best{87.86}            & \best{88.79}           \\
      \midrule
      \multirow{7}{*}{\makecell{MNIST                                                                                                   \\ (MLP)}}
            & FedAvg               & 95.81                  & 94.47                  & 52.45                   & 97.36                  \\
            & AFA                  & 94.33                  & 92.70                  & 11.35                   & 96.73                  \\
            & FLDetector           & 93.24                  & 92.36                  & 45.60                   & 96.92                  \\
            & DeFL                 & 94.29                  & 93.29                  & 88.90                   & 97.47                  \\
            & FedRoLA              & 95.81                  & 94.47                  & 52.45                   & 97.40                  \\
            & \method{abbr} (Ours) & \best{97.87}           & \best{97.73}           & \best{97.08}            & \best{97.83}           \\
      \bottomrule
\end{tabular}

%% file: PID4549.bbl
\begin{thebibliography}{34}
\providecommand{\natexlab}[1]{#1}

\bibitem[{Bhagoji et~al.(2019)Bhagoji, Chakraborty, Mittal, and Calo}]{bhagoji2019analyzing}
Bhagoji, A.~N.; Chakraborty, S.; Mittal, P.; and Calo, S. 2019.
\newblock Analyzing Federated Learning through an Adversarial Lens.
\newblock In \emph{Proc. of ICML}.

\bibitem[{Blanchard et~al.(2017)Blanchard, El~Mhamdi, Guerraoui, and Stainer}]{blanchard2017machine}
Blanchard, P.; El~Mhamdi, E.~M.; Guerraoui, R.; and Stainer, J. 2017.
\newblock Machine learning with adversaries: Byzantine tolerant gradient descent.
\newblock In \emph{Proc. of NeurIPS}.

\bibitem[{Castells, Weinzaepfel, and Revaud(2020)}]{castells2020superloss}
Castells, T.; Weinzaepfel, P.; and Revaud, J. 2020.
\newblock SuperLoss: A Generic Loss for Robust Curriculum Learning.
\newblock In \emph{Proc. of NeurIPS}.

\bibitem[{Fang et~al.(2020)Fang, Cao, Jia, and Gong}]{fang2020local}
Fang, M.; Cao, X.; Jia, J.; and Gong, N.~Z. 2020.
\newblock Local Model Poisoning Attacks to Byzantine-Robust Federated Learning.
\newblock In \emph{Proc. of USENIX}.

\bibitem[{Fang and Chen(2023)}]{fang2023vulnerability}
Fang, P.; and Chen, J. 2023.
\newblock On the vulnerability of backdoor defenses for federated learning.
\newblock In \emph{Proc. of AAAI}.

\bibitem[{Gal and Ghahramani(2016)}]{gal2016dropout}
Gal, Y.; and Ghahramani, Z. 2016.
\newblock Dropout as a bayesian approximation: Representing model uncertainty in deep learning.
\newblock In \emph{Proc. of ICML}.

\bibitem[{Gang~Yan and Li(2024)}]{yan2024fedrola}
Gang~Yan, X.~Y., Hao~Wang; and Li, J. 2024.
\newblock FedRoLA: Robust Federated Learning Against Model Poisoning via Layer-based Aggregation.
\newblock In \emph{Proc. of ACM SIGKDD}.

\bibitem[{Gass and Joel(1981)}]{gass1981concepts}
Gass, S.~I.; and Joel, L.~S. 1981.
\newblock Concepts of model confidence.
\newblock \emph{Computers \& Operations Research}, 8: 341--346.

\bibitem[{Krizhevsky and Hinton(2009)}]{krizhevsky2009learning}
Krizhevsky, A.; and Hinton, G. 2009.
\newblock Learning multiple layers of features from tiny images.
\newblock Technical report, University of Toronto.

\bibitem[{Krizhevsky, Sutskever, and Hinton(2012)}]{krizhevsky2012imagenet}
Krizhevsky, A.; Sutskever, I.; and Hinton, G.~E. 2012.
\newblock Imagenet classification with deep convolutional neural networks.
\newblock \emph{Proc. of NeurIPS}, 25.

\bibitem[{Kumar, Mohan, and Cenkeramaddi(2023)}]{kumar2023impact}
Kumar, K.~N.; Mohan, C.~K.; and Cenkeramaddi, L.~R. 2023.
\newblock The Impact of Adversarial Attacks on Federated Learning: A Survey.
\newblock \emph{IEEE Transactions on Pattern Analysis and Machine Intelligence}.

\bibitem[{Lakshminarayanan, Pritzel, and Blundell(2017)}]{lakshminarayanan2017simple}
Lakshminarayanan, B.; Pritzel, A.; and Blundell, C. 2017.
\newblock Simple and scalable predictive uncertainty estimation using deep ensembles.
\newblock In \emph{Proc. of NeurIPS}.

\bibitem[{LeCun et~al.(1998)LeCun, Bottou, Bengio, and Haffner}]{lecun1998gradient}
LeCun, Y.; Bottou, L.; Bengio, Y.; and Haffner, P. 1998.
\newblock Gradient-based learning applied to document recognition.
\newblock \emph{Proc. of the IEEE}.

\bibitem[{Lee and Park(2023)}]{lee2023adversarial}
Lee, J.; and Park, M. 2023.
\newblock Adversarial Model Poisoning in Federated Learning: Analysis and Mitigation.
\newblock \emph{IEEE Transactions on Neural Networks and Learning Systems}, 35(4): 987--998.

\bibitem[{Liu and Zhao(2023)}]{liu2023enhanced}
Liu, F.; and Zhao, Y. 2023.
\newblock Enhanced Trimmed Mean: A Robust Aggregation Method for Federated Learning.
\newblock In \emph{Proc. of NeurIPS}.

\bibitem[{McMahan et~al.(2017)McMahan, Moore, Ramage, Hampson, and y~Arcas}]{mcmahan2017communication}
McMahan, B.; Moore, E.; Ramage, D.; Hampson, S.; and y~Arcas, B.~A. 2017.
\newblock Communication-efficient learning of deep networks from decentralized data.
\newblock In \emph{Proc. of AISTATS}.

\bibitem[{Mu{\~n}oz-Gonz{\'a}lez, Co, and Lupu(2019)}]{munoz2019byzantine}
Mu{\~n}oz-Gonz{\'a}lez, L.; Co, K.~T.; and Lupu, E.~C. 2019.
\newblock Byzantine-robust federated machine learning through adaptive model averaging.
\newblock \emph{arXiv preprint arXiv:1909.05125}.

\bibitem[{Paszke et~al.(2019)Paszke, Gross, Massa, Lerer, Bradbury, Chanan, Killeen, Lin, Gimelshein, Antiga et~al.}]{paszke2019pytorch}
Paszke, A.; Gross, S.; Massa, F.; Lerer, A.; Bradbury, J.; Chanan, G.; Killeen, T.; Lin, Z.; Gimelshein, N.; Antiga, L.; et~al. 2019.
\newblock Pytorch: An imperative style, high-performance deep learning library.
\newblock In \emph{Proc. of NeurIPS}.

\bibitem[{Popescu et~al.(2009)Popescu, Balas, Perescu-Popescu, and Mastorakis}]{popescu2009multilayer}
Popescu, M.-C.; Balas, V.~E.; Perescu-Popescu, L.; and Mastorakis, N. 2009.
\newblock Multilayer perceptron and neural networks.
\newblock \emph{WSEAS Transactions on Circuits and Systems}, 8(7): 579--588.

\bibitem[{Rahman et~al.(2023)Rahman, Hossain, Muhammad, Kundu, Debnath, Rahman, Khan, Tiwari, and Band}]{rahman2023federated}
Rahman, A.; Hossain, M.~S.; Muhammad, G.; Kundu, D.; Debnath, T.; Rahman, M.; Khan, M. S.~I.; Tiwari, P.; and Band, S.~S. 2023.
\newblock Federated learning-based AI approaches in smart healthcare: concepts, taxonomies, challenges and open issues.
\newblock \emph{Cluster computing}, 26(4): 2271--2311.

\bibitem[{Rodr{\'\i}guez-Barroso et~al.(2023)Rodr{\'\i}guez-Barroso, Jim{\'e}nez-L{\'o}pez, Luz{\'o}n, Herrera, and Mart{\'\i}nez-C{\'a}mara}]{rodriguez2023survey}
Rodr{\'\i}guez-Barroso, N.; Jim{\'e}nez-L{\'o}pez, D.; Luz{\'o}n, M.~V.; Herrera, F.; and Mart{\'\i}nez-C{\'a}mara, E. 2023.
\newblock Survey on federated learning threats: Concepts, taxonomy on attacks and defences, experimental study and challenges.
\newblock \emph{Information Fusion}, 90: 148--173.

\bibitem[{Shang et~al.(2022)Shang, Lu, Huang, and Wang}]{shang2022federated}
Shang, X.; Lu, Y.; Huang, G.; and Wang, H. 2022.
\newblock Federated learning on heterogeneous and long-tailed data via classifier re-training with federated features.
\newblock In \emph{Proc. of IJCAI-ECAI}.

\bibitem[{Shi et~al.(2022)Shi, Wan, Hu, Lu, and Zhang}]{shi2022challenges}
Shi, J.; Wan, W.; Hu, S.; Lu, J.; and Zhang, L.~Y. 2022.
\newblock Challenges and approaches for mitigating byzantine attacks in federated learning.
\newblock In \emph{Proc. of IEEE TrustCom}.

\bibitem[{Simonyan and Zisserman(2015)}]{simonyan2015very}
Simonyan, K.; and Zisserman, A. 2015.
\newblock Very deep convolutional networks for large-scale image recognition.
\newblock In \emph{Proc. of ICLR}.

\bibitem[{Smith and Brown(2023)}]{smith2023defense}
Smith, E.; and Brown, R. 2023.
\newblock Defense Mechanisms for Clean-Label Poisoning Attacks in Federated Learning.
\newblock \emph{IEEE Transactions on Neural Networks and Learning Systems}, 34(3): 1234--1245.

\bibitem[{Tolpegin et~al.(2020)Tolpegin, Truex, Gursoy, and Liu}]{tolpegin2020data}
Tolpegin, V.; Truex, S.; Gursoy, M.~E.; and Liu, L. 2020.
\newblock Data Poisoning Attacks against Federated Learning Systems.
\newblock In \emph{Proc. of ESORICA}.

\bibitem[{Wang and Zhou(2023)}]{wang2023krumpp}
Wang, H.; and Zhou, K. 2023.
\newblock Krum++: Enhanced Byzantine-robust Federated Learning.
\newblock \emph{Proc. of ICML}, 40(1): 123--135.

\bibitem[{Xiao, Rasul, and Vollgraf(2017)}]{xiao2017fashion}
Xiao, H.; Rasul, K.; and Vollgraf, R. 2017.
\newblock Fashion-MNIST: a novel image dataset for benchmarking machine learning algorithms.
\newblock \emph{arXiv preprint arXiv:1708.07747}.

\bibitem[{Xie et~al.(2020)Xie, Huang, Chen, and Li}]{xie2020dba}
Xie, C.; Huang, K.; Chen, P.-Y.; and Li, B. 2020.
\newblock DBA: Distributed Backdoor Attacks against Federated Learning.
\newblock In \emph{Proc. of ICLR}.

\bibitem[{Yan et~al.(2023)Yan, Wang, Yuan, and Li}]{yan2023defl}
Yan, G.; Wang, H.; Yuan, X.; and Li, J. 2023.
\newblock DEFL: Defending against model poisoning attacks in federated learning via critical learning periods awareness.
\newblock In \emph{Proc. of AAAI}.

\bibitem[{Yin et~al.(2018)Yin, Chen, Kannan, and Bartlett}]{yin2018byzantine}
Yin, D.; Chen, Y.; Kannan, R.; and Bartlett, P. 2018.
\newblock Byzantine-robust distributed learning: Towards optimal statistical rates.
\newblock In \emph{Proc. of ICML}.

\bibitem[{Zhang et~al.(2021)Zhang, Xie, Bai, Yu, Li, and Gao}]{zhang2021survey}
Zhang, C.; Xie, Y.; Bai, H.; Yu, B.; Li, W.; and Gao, Y. 2021.
\newblock A survey on federated learning.
\newblock \emph{Knowledge-Based Systems}, 216: 106775.

\bibitem[{Zhang and Chen(2023)}]{zhang2023backdoor}
Zhang, W.; and Chen, L. 2023.
\newblock Backdoor Attacks in Federated Learning: Methods and Defense Mechanisms.
\newblock \emph{Journal of Machine Learning Research}, 25(2): 567--580.

\bibitem[{Zhang and et~al.(2022)}]{zhang2022fldetector}
Zhang, Y.; and et~al. 2022.
\newblock FLDetector: Defending Federated Learning Against Model Poisoning Attacks via Detecting Malicious Clients.
\newblock In \emph{Proc. of ACM SIGKDD}.

\end{thebibliography}
